\documentclass[letterpaper]{article} % DO NOT CHANGE THIS
\usepackage{aaai2026}  % DO NOT CHANGE THIS
\usepackage{times}  % DO NOT CHANGE THIS
\usepackage{helvet}  % DO NOT CHANGE THIS
\usepackage{courier}  % DO NOT CHANGE THIS
\usepackage[hyphens]{url}  % DO NOT CHANGE THIS
\usepackage{graphicx} % DO NOT CHANGE THIS
\usepackage{natbib}  % DO NOT CHANGE THIS AND DO NOT ADD ANY OPTIONS TO IT
\usepackage{caption} % DO NOT CHANGE THIS AND DO NOT ADD ANY OPTIONS TO IT
\usepackage{algorithm}
\usepackage{algorithmic}
\usepackage{multirow} 
\usepackage{amsfonts}
\usepackage[table]{xcolor}
\usepackage{array}

\usepackage{newfloat}
\usepackage{listings}
\DeclareCaptionStyle{ruled}{labelfont=normalfont,labelsep=colon,strut=off} % DO NOT CHANGE THIS
\floatstyle{ruled}
\newfloat{listing}{tb}{lst}{}
\floatname{listing}{Listing}
\usepackage{amsmath}

\title{Activating Visual Context and Commonsense Reasoning \\ through Masked Prediction in VLMs}
\author{
    Jiaao Yu, Shenwei Li, Mingjie Han, Yifei Yin, Wenzheng Song, Chenghao Jia, Man Lan
}
\affiliations{
    \textsuperscript{\rm 1}
    School of Computer Science and Technology,
    East China Normal University
}

\usepackage{bibentry}

\begin{document}

\maketitle

\begin{abstract}

Recent breakthroughs in reasoning models have markedly advanced the reasoning capabilities of large language models, particularly via training on tasks with verifiable rewards. Yet, a significant gap persists in their adaptation to real-world multimodal scenarios, most notably, vision-language tasks, due to a heavy focus on single-modal language settings. While efforts to transplant reinforcement learning techniques from NLP to Visual Language Models (VLMs) have emerged, these approaches often remain confined to perception-centric tasks or reduce images to textual summaries, failing to fully exploit visual context and commonsense knowledge, ultimately constraining the generalization of reasoning capabilities across diverse multimodal environments. To address this limitation, we introduce a novel fine-tuning task, Masked Prediction via Context and Commonsense (MPCC), which forces models to integrate visual context and commonsense reasoning by reconstructing semantically meaningful content from occluded images, thereby laying the foundation for generalized reasoning. To systematically evaluate the model’s performance in generalized reasoning, we developed a specialized evaluation benchmark, MPCC-Eval, and employed various fine-tuning strategies to guide reasoning. Among these, we introduced an innovative training method, Reinforcement Fine-Tuning with Prior Sampling, which not only enhances model performance but also improves its generalized reasoning capabilities in out-of-distribution (OOD) and cross-task scenarios. Code and data are available at \url{yjainqdc.github.io/MPCC_}.

\end{abstract}

\begin{figure}[t]
\centering
\includegraphics[width=0.46\textwidth]{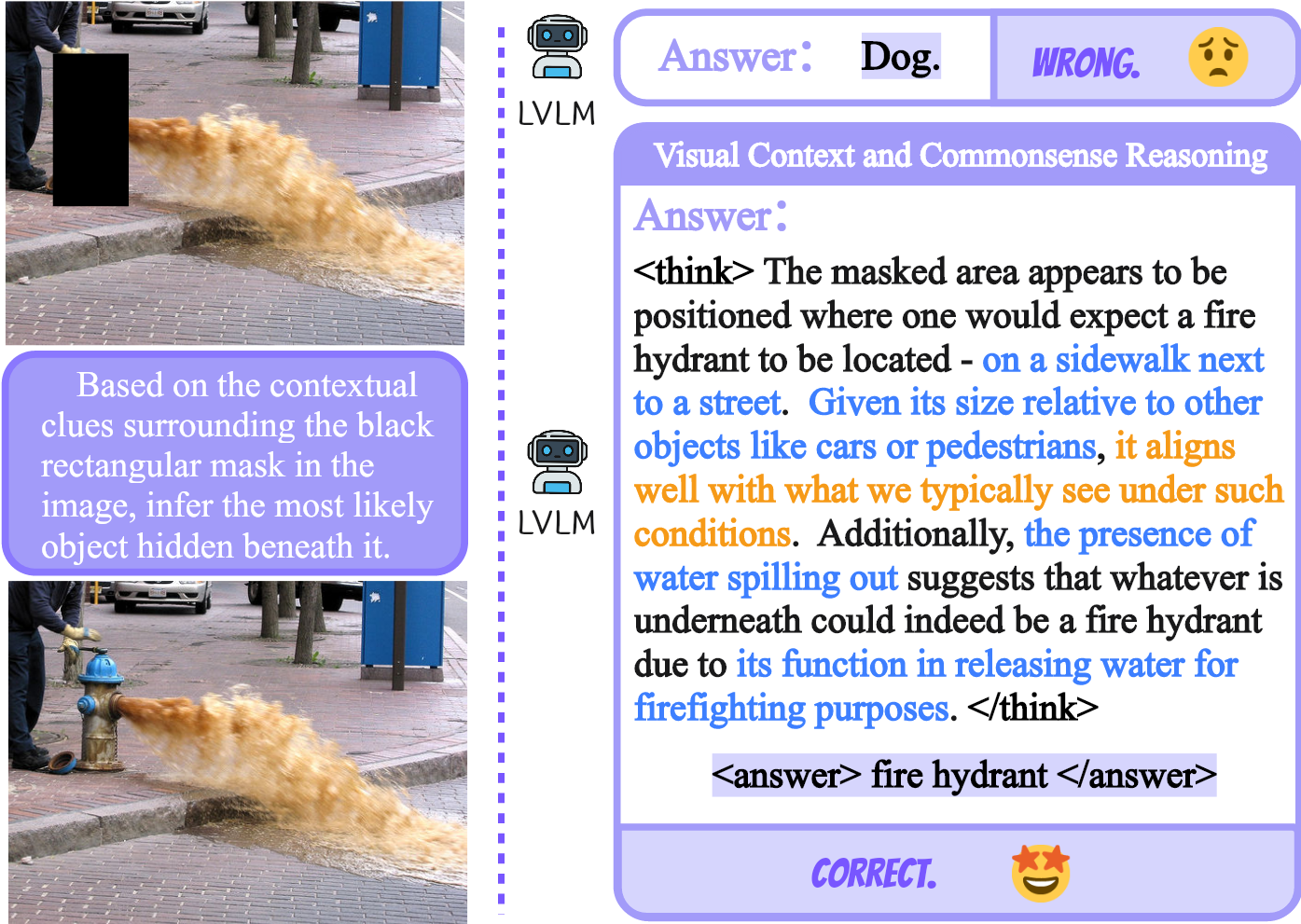}
%The example of Masked Prediction via Context and Commonsense (MPCC) task. Qwen2.5VL-3B回答了错误答案，经过RFT with 先验采样的模型推理并回答了正确答案。
\caption{The example of Masked Prediction via Context and Commonsense (MPCC) task. In this example, we highlighted the visual context reasoning part in blue and the commonsense reasoning part in orange.}
\label{fig1}
\end{figure}

\begin{figure*}[t]
\centering
\includegraphics[width=1.0\textwidth]{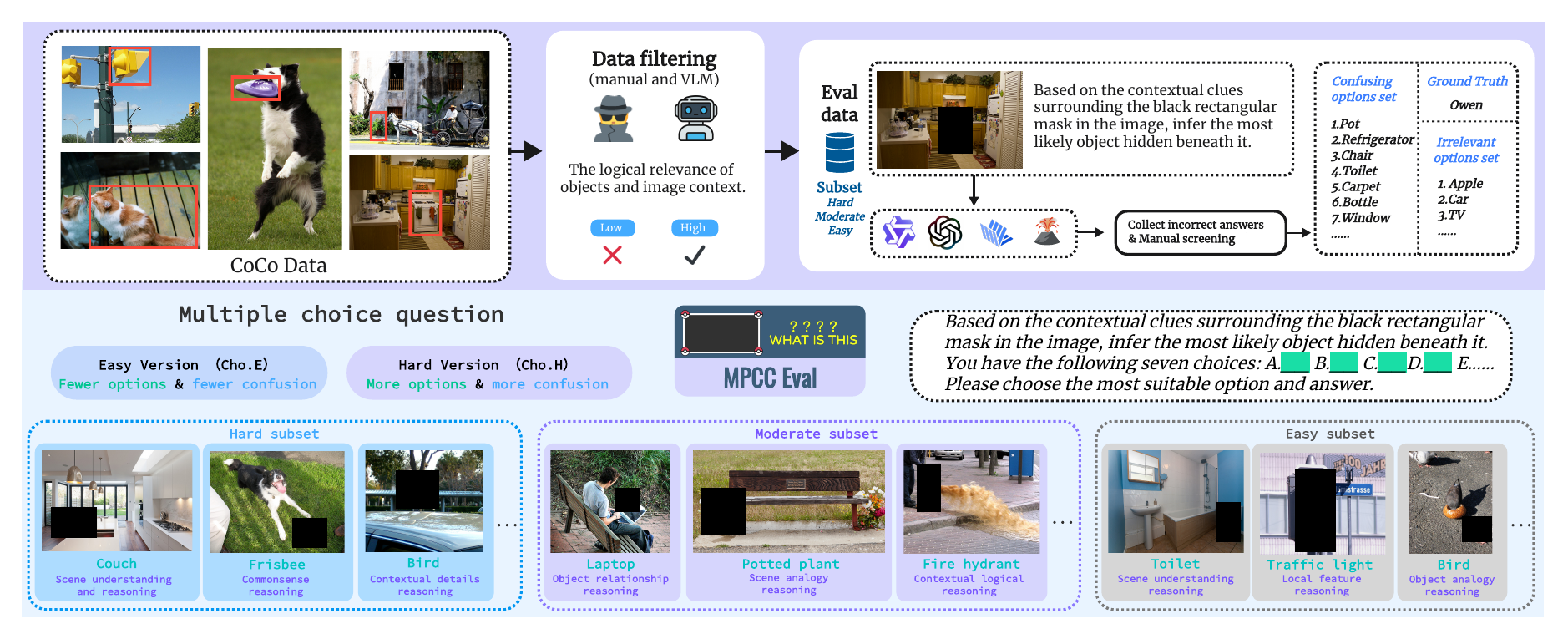} % Reduce the figure size so that it is slightly narrower than the column.
\caption{The MPCC-Eval benchmark creation pipeline filters and curates images by difficulty, forming three subsets: easy, moderate, and hard. It includes two types of single-choice question formats.}
\label{fig2}
\end{figure*}

\section{Introduction}

Recent advances in reasoning models, such as DeepSeek-R1 \cite{guo2025deepseek} and OpenAI o1 \cite{jaech2024openai}, have significantly enhanced the reasoning capabilities of large language models. Notably, DeepSeek-R1 introduces a simple yet effective rule-based reinforcement learning (RL) framework \cite{shao2024deepseekmath}, which enables emergent reasoning behavior without relying on traditional techniques such as Monte Carlo Tree Search  \cite{xin2024deepseek,gao2024interpretable} or Process Reward Models (PRM) \cite{lightman2023let}. This development has introduced new possibilities in post-training methodologies, inspiring the design of more powerful reasoning architectures \cite{wang2025beyond,hu2025open}. Nonetheless, most existing research remains focused on single-modal language settings. Visual reasoning, in particular, plays a crucial role in understanding complex, multimodal data, and is essential for both advancing domain-specific applications and for progressing toward Artificial General Intelligence. Despite recent progress, improving and evaluating reasoning capabilities in multimodal large models remains an open and challenging research problem.

DeepSeek-R1 achieves generalization from structured tasks to natural language by using code generation and mathematical reasoning that tasks with verifiable rewards as training "anchors." The alignment in objectives between these structured and language tasks enables effective cross-task generalization. However, unlike NLP, visual reasoning lacks explicit, verifiable reward signals. Although recent efforts have adapted RL techniques from NLP to boost the reasoning abilities of multimodal large models such as Visual-RFT \cite{liu2025visual}, PR1 \cite{yu2025perception}, Vlm-r1 \cite{shen2025vlm} and Visual-R1 \cite{huang2025vision}, these approaches are mostly limited to perception tasks or reduce images to text, falling short of deep reasoning over visual context and inherent commonsense of large language models. Due to heterogeneity, perception tasks such as detection struggle to generalize to broader reasoning scenarios.
Such approaches fail to fully exploit the rich visual context and commonsense knowledge inherent in language models, ultimately hindering the development of robust generalized reasoning abilities.
To address these limitations, we explore a simple yet effective task design to efficiently activate and evaluate visual context and commonsense reasoning in multimodal large models, and investigate practical fine-tuning strategies to enhance their multimodal reasoning abilities through empirical study.

In conventional deep learning, Masked Prediction is a simple yet effective approach to improve the representational capacity of a model and improve generalization. By randomly masking local information from input data, such as tokens in text or patches in images, the model is compelled to reconstruct the missing content based on contextual cues. This training paradigm imposes inference pressure in the presence of incomplete information, encouraging the model to learn the underlying structure of the data and capture cross-dimensional associations, thereby improving the robustness and generalizability of its representations. Inspired by prior work, we propose the Masked Prediction via Context and Commonsense (MPCC) task. Specifically, MPCC masks key visual entities to force the model to integrate commonsense knowledge with visual contextual cues, enabling cross-modal compensatory reasoning. This approach aims to activate the visual context understanding and commonsense reasoning capabilities of visual language models (VLMs), enabling them to predict the masked objects in images accurately.
As shown in Figure \ref{fig1}, the fire hydrant is covered, but large models often struggle to perform visual context and commonsense reasoning, leading to incorrect answers. After incorporating contextual and commonsense reasoning, the model provided the correct answer.

Building on this, we introduce a dedicated evaluation benchmark to assess the proposed capability in large models. The benchmark includes 1,114 images with answers and confusing items.  Tasks are single-choice and grouped by difficulty, enabling comprehensive evaluation of model's abilities in visual context and commonsense reasoning under varying complexity.

Recent studies have focused on activating reasoning in models through fine-tuning with limited data. Prompting methods guide models to reason before answering, while supervised fine-tuning (SFT) on large language models enhances reasoning via end-to-end training. However, SFT relies heavily on high-quality CoT annotations and complex data strategies, often leading to overfitting and poor generalization. Reinforcement fine-tuning (RFT), especially with GRPO \cite{shao2024deepseekmath}, has shown promise in boosting performance and reasoning ability through reward shaping. Some works have explored combining SFT and RFT in two-stage post-training \cite{tan2025reason} on vision language models. On the MPCC task, we evaluate the impact of different fine-tuning approaches on reasoning, focusing on performance, out-of-distribution (OOD) generalization, and cross-task generalization. In practice, post-training data often include verified outcomes and partial reasoning labels. We propose a priori sampling within RFT using these labels to guide initial policy generation, combining human-labeled and model-generated samples for advantage estimation and policy updates. This serves as a preliminary exploration toward better performance-generalization trade-offs.

\begin{figure*}[t]
\centering
\includegraphics[width=1.0\textwidth]{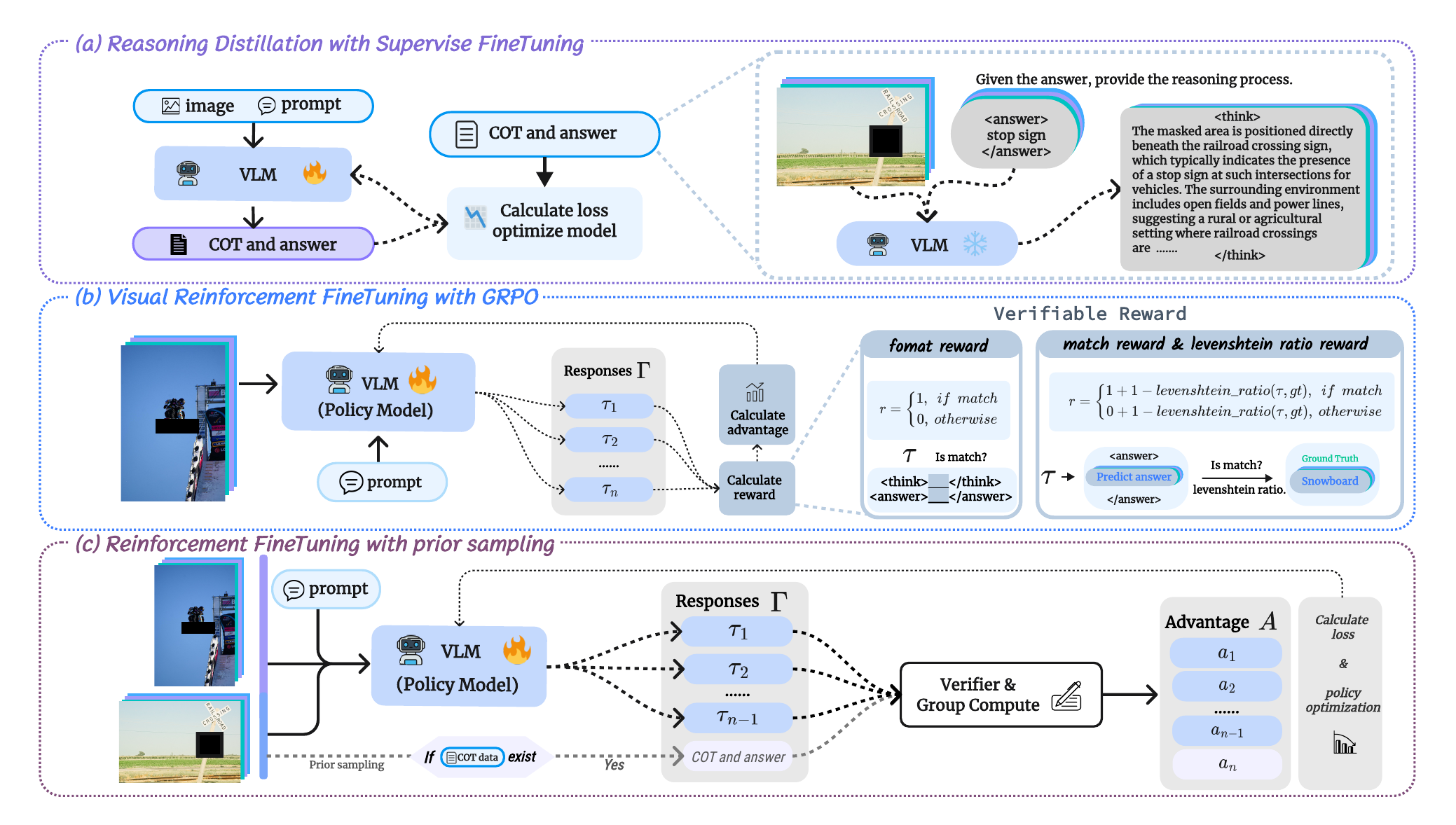} % Reduce the figure size so that it is slightly narrower than the column.
\caption{Fine-tuning Strategies on MPCC Task. (a) SFT: Construct and distill reasoning data via supervised fine-tuning. (b) GRPO-based RFT: Activate reasoning with verifiable rewards. (c) Proposed RFT with Prior Sampling: Use annotated reasoning data to replace one sampling step.}
\label{fig3}
\end{figure*}

% 3.我们研究了不同推理引导训练策略对多模态大型模型性能和泛化能力的影响，并提出RFT with 先验采样，探索如何有效利用有限数据来激活上下文和常识推理能力。
The main contributions are summarized as follows:

\begin{itemize}
    \item We propose a fine-tuning task, Masked Prediction via Context and Commonsense (MPCC), which activates visual context and commonsense reasoning through masked prediction, thereby enhancing the generalized reasoning ability of VLMs.
    \item We explore the impact of reasoning-guided fine-tuning strategies on the performance and generalization of VLMs. Additionally, we propose RFT with prior sampling to effectively activate reasoning.
    \item We contribute MPCC-Eval, a benchmark designed to assess multimodal models' ability to perform masked prediction with context and commonsense awareness.
\end{itemize}

\section{Related Work}

\subsection{Visual reasoning}

Visual Language Models (VLMs) such as LLaVA \cite{liu2023visual}, Qwen-VL \cite{wang2024qwen2,bai2025qwen2}, InternVL \cite{chen2024internvl}, and GPT-4o \cite{hurst2024gpt} have driven rapid progress in multimodal understanding and generation, yet their visual reasoning, particularly in visual context and commonsense, remains limited. 
Early advances in reasoning activation primarily leveraged prompt engineering techniques, such as Chain-of-Thought (CoT) \cite{wei2022chain} and tree of thought \cite{yao2023tree}. In the vision-language domain, methods like LLaVA-CoT \cite{xu2024llava} utilized multi-stage supervised fine-tuning (SFT) with CoT supervision, while Insight-V \cite{dong2025insight} integrated SFT with reinforcement learning. 
The recent success of reasoning-centric models like DeepSeek-R1 \cite{guo2025deepseek} has shifted the focus of large language model (LLM) research toward post-training with RL \cite{zuo2025ttrl,zhao2025echo} and reasoning-driven capability enhancement. Notably, GRPO \cite{shao2024deepseekmath} has demonstrated that robust reasoning abilities can be induced through RL with verifiable rewards, effectively eliminating the need for SFT.
Despite these advances, RL-based reasoning methods have been largely confined to the language domain.
Recent efforts to address these shortcomings have primarily adopted language-centric strategies, formalizing visual structures linguistically and improving reasoning through supervised or reinforcement learning, as in R1-OneVision \cite{yang2025r1}, Vision-R1 \cite{huang2025vision}, and LMM-R1 \cite{peng2025lmm}. However, these approaches largely target mathematical reasoning and thus lack strong generalization. Alternatively, some studies directly apply RL to visual tasks using verifiable rewards (e.g., IoU, character matching) to activate reasoning for perception tasks like detection and counting \cite{liu2025visual,yu2025perception,chen2025g1}, or adopt multi-stage reasoning fine-tuning \cite{tan2025reason}, but these remain focused on perceptual or mathematical challenges, falling short of open-world visual context and commonsense reasoning. In this work, we systematically compared the performance and generalization of multiple reasoning-guided strategies.

\subsection{Mask Prediction task}

The core idea of masked prediction is inspired by Gestalt psychology, which suggests that the human brain can infer missing elements based on contextual information. In deep learning, this is achieved by randomly masking parts of the input \cite{yang2024polymax,zhang2024pcp,ghazvininejad2019mask}, such as image patches \cite{xie2022simmim,wang2023hard} or text tokens \cite{ghazvininejad2019mask,kaneko2022unmasking}, forcing the model to reconstruct missing content from context, thereby learning intrinsic data structures and relationships.
In NLP, models like BERT \cite{koroteev2021bert} apply dynamic masking to input tokens to capture bidirectional context. In computer vision, MAE \cite{he2022masked} adopts a similar approach by masking visual tokens to strengthen visual representations.
Building on these insights, we treat image semantics as holistic information comprising interrelated elements. By masking key components of images, we encourage models to utilize contextual cues for inference and reasoning.

\section{Methodology}
\subsection{Task definition}

In this work, we propose the Mask Prediction via Context and Commonsense (MPCC) task, which aims to activate VLMs in visual context and commonsense reasoning, thereby enhancing the generalized reasoning ability of VLMs. By masking key visual entities, the model is compelled to integrate commonsense knowledge with visual contextual cues, enabling cross-modal compensatory reasoning and accurate prediction of masked objects within an image.
Given an image $I$, we denote the masked region as $m$. The model is expected to infer and predict the identity of the object $o$ corresponding to $m$ based on reasoning over the available context. Specifically, we input the masked image $I-m$ along with a prompt $p$ into a VLM, obtaining a response $y$. The output $y$ is divided into two parts: the reasoning process $y_r$ and the final answer. Ultimately, our goal is to maximize the probability that the model will answer correctly. The probability is calculated by Eq.\ref{eq:eq1}.

\begin{equation}
    p(y\mid x) = p(y_r \mid I-m,p) \cdot p(o\mid I-m,p,y_r)
  \label{eq:eq1}
\end{equation}

\subsection{Benchmark and Data}

Based on this task, we construct a multimodal contextual and commonsense reasoning benchmark, MPCC-Eval, to evaluate model performance on the MPCC task. The construction pipeline of MPCC-Eval is illustrated in Figure \ref{fig3}. The benchmark is built upon the COCO \cite{lin2014microsoft} validation dataset. First, we employ a visual language model to perform preliminary analysis on the bboxed objects in COCO, assessing the strength of their logical relationships within the image context. Samples with weak contextual associations are filtered out. Based on this filtering, we further conduct manual curation to construct a high-quality test set consisting of 1,114 mask-image pairs. These samples are categorized into three difficulty levels, easy, moderate, and hard according to the complexity of the required reasoning.

As shown in Figure \ref{fig2}, the hard subset primarily involves reasoning tasks such as scene understanding and contextual details; the moderate subset includes tasks mainly centered around object relationship and scene analogy; whereas the easy subset focuses on simpler forms of reasoning, including scene understanding, recognition based on local features, and object analogy.
We further utilize model-generated incorrect answers to construct confusing items. By aggregating and analyzing erroneous responses from multiple large models, we generate 4 confusing items and 2 irrelevant items for each mask-image pair. We format the evaluation as a single choice task, with two difficulty settings: easy and hard.
The easy version includes ground truth (GT), one irrelevant item, and two confusing items, totaling four choices. The hard version contains the GT, two irrelevant items, and four confusing items, resulting in seven candidate options.

For training data, we adopted a similar screening method. For a part of the data, as shown in Figure \ref{fig3} (a), we provide the model with the masked image $I-m$ and the final answer $o$, and allow the model to simulate the thought process. In this way, we constructed SFT data containing reasoning data.

 \subsection{Guide Reasoning}
This section presents the various reasoning-guided training strategies we implemented for the MPCC task, including prompt-based, SFT, and RFT approaches. Additionally, we further apply GRPO-based RFT with prior sampling, leveraging partially labeled reasoning traces.

\textbf{Prompt Guides Reasoning.} 
The simplest way to guide reasoning is prompt engineering, which uses prompt to guide the model to reason first and then answer. The specific system prompt in our work is as Table \ref{table0}:

    \begin{table}[htbp]
    \normalsize
    \centering
    % prompt
    \renewcommand{\arraystretch}{1.2}
    \scalebox{0.88}{
    \begin{tabular}{p{9cm}}
    \hline
        \textbf{System prompt:} \\
        A conversation between User and Assistant. The user asks a question, and the Assistant solves it with think and answer. The assistant first thinks about the reasoning process in the mind and then provides the user with the answer. The reasoning process and answer are enclosed within $<think> </think>$ and $<answer> </answer>$ tags, respectively, i.e.,"$<think>$ reasoning process here $</think><answer>$ answer here $</answer>$. \\
    \hline
    \end{tabular}}
    \caption{System prompt that guide reasoning.}
    \label{table0}
    \end{table}

\textbf{SFT Guides Reasoning.} 
Based on the train dataset constructed above, we further employ Supervised Fine-Tuning (SFT) to guide the model in learning the reasoning process before generating answers as shown in Figure \ref{fig3} (a). During training, we supervise both the thinking and answering phases at the token level, and optimize the model using the standard language modeling loss as Eq.\ref{eq:eq2}, $y$ represents the sequence of responses that includes the reasoning process, and $N$ represents its length.

\begin{equation}
    \mathcal{L}_{\text{SFT}}(\theta) = -\frac{1}{N} \sum_{i=1}^{N} \log \pi_{\theta}(y_i \mid I-m,p,y_{<i})
    \label{eq:eq2}
\end{equation}

\textbf{RFT Guides Reasoning.} 

The DeepSeek-R1 algorithm is trained using a reinforcement learning framework that effectively removes the reliance on supervised fine-tuning. Central to this approach is the Group Relative Policy Optimization (GRPO) framework. In contrast to PPO \cite{schulman2017proximal}, which depend on a separate critic model to assess policy performance, GRPO operates without the need for an external reward model. As Figure \ref{fig3} (b), it enables policy updates by performing direct comparisons among multiple candidate responses. In this process, the candidate responses are evaluated using the following reward function $R$ in Eq.\ref{eq:eq3}:

\begin{equation}
    R(a,o) = r_{fmt} + 1_{\{a = o\}} + 1 - levenshtein(a,o)
  \label{eq:eq3}
\end{equation}

In this setup, $a$ denotes the answer generated after reasoning, and $o$ is the ground-truth answer. The term $1_{\{\cdot\}}$ represents an indicator function that takes the value 1 when the condition inside the brackets is satisfied, and 0 otherwise. The format reward, denoted as $r_{fmt}$, is defined as $1_{\{a\ matches\ format\}}$, indicating whether the generated answer adheres to the required structure. Additionally, we incorporate a complementary reward component based on the levenshtein ratio between the generated answer $a$ and the correct answer $o$. Finally, the GRPO framework is employed to perform group-wise optimization over the candidate responses. The objective of GRPO can be formally defined as maximizing the expected relative log-probability of responses weighted by their normalized rewards over the question distribution and policy sampling : 

\begin{equation}
    \max_{\theta} \mathbb{E}_{q, \{\tau_i\}} \left[ 
        \sum_{i=1}^{G} \frac{r_i - \mu}{\sigma} \cdot \log \frac{\pi_\theta(\tau_i | q)}{\pi_{\text{ref}}(\tau_i | q)}
    \right]
    \label{eq:eq4}
\end{equation}

In Eq.\ref{eq:eq4}, the current policy $\pi_\theta$ generates $G$ candidate responses $\{\tau_i\}$ for a given input, and their corresponding rewards $\{r_i\}$ are computed. $\pi_\text{ref}$ denotes the reference policy. These rewards are normalized using the group mean $\mu$ and standard deviation $\sigma$ to reflect each response’s relative advantage. We perform reinforcement fine-tuning based on GRPO using verifiability rewards, enabling the training of reasoning capabilities even when only final answers are available as supervision.

In this work, we apply prompt engineering, supervised fine-tuning, and reinforcement fine-tuning to our proposed MPCC task for enhancing reasoning capabilities. We evaluate the performance of each individual method as well as their combinations, such as RFT after SFT.

\textbf{RFT with prior sampling.} 
In practical post-training scenarios, most publicly available datasets are in the form of query-answer pairs, while query-think-answer (i.e., explicit reasoning) data are relatively scarce. The former readily supports verifiable reward-based reinforcement learning, whereas the latter is suitable for supervised fine-tuning (SFT). However, when data are limited, verifiable reinforcement fine-tuning tends to be less efficient compared to SFT. Yet, SFT’s token-level supervision may lead to overfitting, resulting in cognitive rigidity and reduced generalization ability.
To balance performance and generalization, we propose a reinforcement fine-tuning approach that integrates both data types. Specifically, we introduce RFT with prior sampling as illustrated Figure \ref{fig3} (c). Specifically, for data with only query-answer pairs, we perform standard RFT. For data annotated with explicit reasoning processes, we replace one model-generated policy sample with the annotated reasoning trajectory. The policy gradient is updated through group optimization that incorporates both model-generated samples and annotated reasoning trajectory. The advantage computation in the group optimization can be formulated as follows:

\begin{equation}
    A_{i} = \frac{r_{i} - \text{mean}(\{r_1, \ldots, r_{G-1}, r_{prior}\})}{\text{std}(\{r_1, \ldots, r_{G-1}, r_{prior}\})}
    \label{eq:eq5}
\end{equation}

In Eq.\ref{eq:eq5}, $r_i$ denotes the reward of the $i$-th sampled trajectory. We compute the advantage $A_i$ for sampled trajectory. For a priori sample, we also leverage its reward $r_{prior}$ calculation advantage $A_{prior}$. %We conducted a detailed comparison of the effectiveness of this method as an exploration of fine-tuning to guide reasoning.

%===

    \begin{table*}[htbp]
    \normalsize
    \centering

    \renewcommand{\arraystretch}{1.0}
    \scalebox{0.9}{
    \begin{tabular}{lccccccccccccc}
    \hline
    Subset & \multicolumn{2}{c}{Hard} && \multicolumn{2}{c}{Moderate} && \multicolumn{2}{c}{Easy} && \multicolumn{2}{c}{Ave.} && Sum \\
    \cline{2-3} \cline{5-6} \cline{8-9} \cline{11-12}

    Model & Cho.E & Cho.H && Cho.E & Cho.H && Cho.E & Cho.H && Cho.E & Cho.H && Score \\
    \hline
        \rowcolor{gray!18}
        Random & 25.00 & 14.29 && 25.00 & 14.29 && 25.00 & 14.29 && 25.00 & 14.29 && 117.87\\
        \rowcolor{gray!12}
        Human & 97.82 & 96.27 && 98.53 & 98.24 && 99.56 & 98.67 && 98.64 & 97.73 && 589.09\\
        Qwen2.5-VL-3B & 48.14 & 21.43 && 63.05 & 27.57 && 66.08 & 41.46 && 59.09 & 30.15 && 267.73\\
        Qwen2.5-VL-7B & 58.39 & 21.12 && 73.02 & 33.43 && 75.83 & 42.79 && 69.08 & 32.45 && 304.58\\
        LLaVA1.6-7B & 41.30 & 18.63 && 48.30 & 22.58 && 47.01 & 17.96 && 45.54 & 19.72 && 195.78\\
        LLaVA1.6-13B & 46.58 & 23.60 && 57.77 & 26.39 && 60.09 & 34.15 && 54.81 & 28.05 && 248.58\\
        Intern3-VL-1B & 44.40 & 25.78 && 49.56 & 32.26 && 58.31 & 34.37 && 50.76 & 30.80 && 244.68\\
        Intern3-VL-8B & 60.56 & 41.61 && 70.67 & 57.18 && 72.72 & 65.41 && 67.98 & 54.73 && 368.15\\
        GPT-4o-mini & 48.76 & 32.61 && 60.70 & 46.63 && 65.19 & 60.09 && 58.22 & 46.34 && 313.98\\
        GPT-4o & 57.45 & 46.58 && 73.31 & 56.01 && 83.15 & 72.50 && 71.30 & 58.36 && 389.00\\

    \hline
    \end{tabular}}
    \caption{The performance of current mainstream vision-language models on the MPCC-Eval benchmark is presented.}
    \label{table1}
    \end{table*}

    \begin{table*}[htbp]
    \normalsize
    \centering
    % BenchMark 实验 -- Judge子集
    \renewcommand{\arraystretch}{1.0}
    \scalebox{0.9}{
    \begin{tabular}{lcccccccccccccccc}
    \hline
    Subset & \multicolumn{4}{c}{Hard} && \multicolumn{4}{c}{Moderate} && \multicolumn{4}{c}{Easy}\\
    \cline{2-5} \cline{7-10} \cline{12-15}
    Model & GT & conf. & irre. & All && GT & conf. & irre. & All && GT & conf. & irre. & All \\
    \hline
        Qwen2.5-VL-3B & 46.27 & 55.82 & 95.19 & 69.01 && 66.57 & 60.56 & 97.80 & 74.25 && 74.28 & 65.96 & 96.23 & 79.33 \\
        Qwen2.5-VL-7B & 36.02 & 70.81 & 98.29 & 74.04 && 53.37 & 74.85 & 99.12 & 79.35 && 69.40 & 75.44 & 99.22 & 83.64 \\
        LLaVA1.6-7B & 48.76 & 53.65 & 86.34 & 65.65 && 60.41 & 55.28 & 89.88 & 66.52 && 69.40 & 58.70 & 89.36 & 71.35 \\
        LLaVA1.6-13B & 56.83 & 45.57 & 79.97 & 62.05 && 69.79 & 48.83 & 84.60 & 65.40 && 73.61 & 53.55 & 84.15 & 68.20\\
        Intern3-VL-1B & 43.79 & 59.86 & 97.21 & 68.26 && 65.85 & 59.02 & 97.11 & 73.31 && 66.96 & 66.35 & 97.67 & 78.67 \\
        Intern3-VL-8B & 48.45 & 65.92 & 96.12 & 73.42 && 67.45 & 68.48 & 99.12 & 79.35 && 75.17 & 70.45 & 98.56 & 81.29\\
        GPT-4o-mini & 39.44 & 62.50 & 96.89 & 68.99 && 58.06 & 66.64 & 99.56 & 74.74 && 74.41 & 84.39 & 99.45 & 85.55\\
        GPT-4o & 69.25 & 54.27 & 94.10 & 71.80 && 78.89 & 60.04 & 97.65 & 80.61 && 87.36 & 77.55 & 98.34 & 88.69 \\

    \hline
    \end{tabular}}
    \caption{Evaluating models' ability to distinguish GT, confusing items (conf.), and irrelevant items (irre.) by true/false questions.}
    \label{table2}
    \end{table*}

\section{Experiments}

\subsection{Experimental Setting}

In the benchmark experiments, we used the official weights of the original models. For closed-source models such as GPT-4o and GPT-4o-mini, we evaluated and reported results via API access. In the reasoning-guided training experiments, we used Qwen2.5-vl-3B as the baseline. The SFT and reinforcement learning algorithms were implemented based on the Transformers \footnote{https://github.com/huggingface/transformers} library and OpenR1 \footnote{https://github.com/huggingface/open-r1}, respectively. All experiments were conducted on two NVIDIA A100-80G GPUs. For detailed settings and experiments, please refer to the appendix.

\subsection{MPCC-Eval Benchmark}

We evaluated the performance of current mainstream models on the MPCC-Eval Benchmark, as shown in Table \ref{table1}. Additionally, we assessed their ability to distinguish between ground truth (GT), confusing items, and irrelevant items, with results presented in Table \ref{table2}.
The results in Table 1 indicate that these models have initially demonstrated capability for this task. However, due to a lack of reasoning skills, they perform poorly on more challenging data. Notably, GPT-4o and models from the InternalVL family achieved superior results. Among them, Intern3-VL-1B and Qwen2.5-VL-3B delivered impressive outcomes despite their smaller model sizes, highlighting their cost-effectiveness. Our reasoning-guided training experiments were conducted using Qwen2.5-VL-3B.
To further analyze the models' capabilities, we tested their ability to differentiate between GT, confusing items, and irrelevant items through true or false questions. The findings revealed distinct answering patterns among different models when faced with difficult problems. For instance, LLaVA tends to answer "Yes" frequently, leading to suboptimal performance in identifying irrelevant items. This suggests that LLaVA may suffer more from hallucination issues compared to other models. Conversely, Qwen2.5-VL-7B often responds with "No," which, while improving its accuracy in detecting irrelevant items, results in errors when identifying GT. These results underscore that current mainstream large language models struggle without proper reasoning abilities, emphasizing the importance of incorporating reasoning into their design.

%--------------------------------------------------------------------------------------------------------------
\begin{figure*}[t]
\centering
\includegraphics[width=0.95\textwidth]{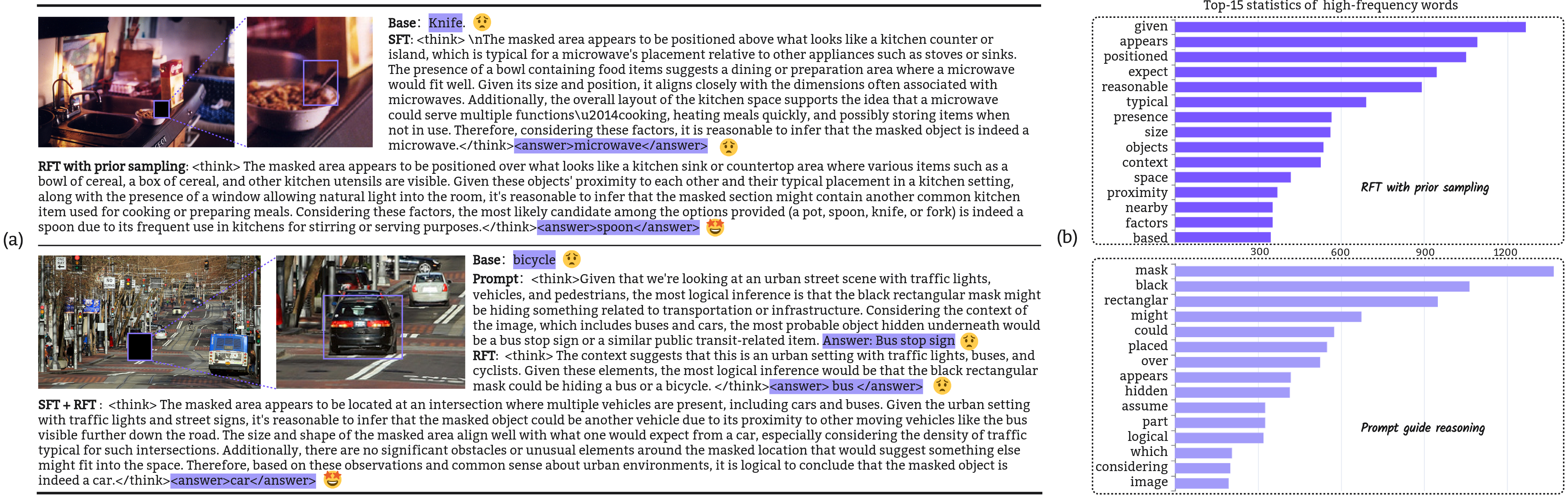}
\caption{Model responses in the MPCC task and high-frequency word statistics after prompt engineering and fine-tuning.}
\label{fig4}
\end{figure*}

    \begin{table}[htbp]
    \normalsize
    \centering
    \renewcommand{\arraystretch}{1.2}
    \scalebox{0.9}{
    \begin{tabular}{lccccccc}
    \hline
    & \multicolumn{2}{c}{In-Distribution(Ave.)} && \multicolumn{2}{c}{Out-of-Distribution} \\
    \cline{2-3} \cline{5-6}
    Methods & Cho.E & Cho.H && Cho.E & Cho.H\\
    \hline
        \rowcolor{gray!25}
        Base & 59.09 & 30.15  && 61.74 & 32.17 \\
        \hspace{1em} -prompt & 59.91 & 33.24 && 64.35 & 43.48 \\
        \hspace{1em} -sft & 74.10 & \textbf{58.85} && 71.88 & 55.65 \\
        \hspace{1em} -rft & 66.42 & 37.18 && \textbf{74.20} & 55.36\\
        \hspace{1em} -sft+rft & \textbf{74.93} & 58.18 && 72.41 & 52.46 \\
        \hspace{1em} -rft+$pri.$  & 72.59 & 51.50 && \textbf{74.20} & \textbf{56.23} \\
    \hline
    \end{tabular}}
    \caption{Comparison of performance and OOD generalization of different fine-tuning strategies.}
    \label{table3}
    \end{table}

\subsection{Fine-tuning Strategies on MPCC}

In this experiments, we evaluated the reasoning capabilities of the MPCC task-activated model through various fine-tuning strategies, including prompt-based methods, supervised fine-tuning (sft), reinforcement fine-tuning (rft), a combination of SFT and RFT (sft+rft), and RFT fine-tuning with prior sampling (rft+$pri.$).

\textbf{In-distribution performance and out-of-distribution (OOD) generalization.} In Table \ref{table3}, we used two settings to evaluate the model’s in-distribution performance and OODgeneralization. The first setting involved training and testing on the full training and evaluation sets, respectively. For the OOD setting, we split the 80 categories of COCO into two disjoint subsets, using one for training and the other for testing. This ensured that the test categories were unseen during training, forming a true OOD scenario and allowing us to assess the model’s generalization under distribution shift.
Under the in-distribution setting, SFT combined with RFT (sft+rft) achieved the best overall performance, yet it performed relatively poorly in the OOD setting. When fine-tuning with in-distribution data, SFT showed more significant performance gains compared to RFT, but it exhibited weaker OOD generalization. Notably, although RFT fine-tuning with prior sampling (rft+$pri.$) under the full dataset setting did not perform as well as SFT+RFT, it demonstrated a good balance in the OOD setting, maintaining a reasonable performance level while showing strong OOD generalization. This suggests that  RFT with prior sampling offers a promising strategy for maintaining robustness under shifts in data distribution.

    \begin{table}[t]
    \normalsize
    \centering
    % BenchMark 实验 -- Judge子集
    \renewcommand{\arraystretch}{1.2}
    \scalebox{0.82}{
    \begin{tabular}{p{7.6em} >{\centering\arraybackslash}m{3.8em} >{\centering\arraybackslash}m{3.8em} >{\centering\arraybackslash}m{3.8em} >{\centering\arraybackslash}m{3.8em}}
    \hline
    Methods & Sem. & Rel. & Obj. & Spa.\\
    \hline
    \multicolumn{5}{c}{\textbf{Fine tune on perceive task}} \\
        \rowcolor{gray!25}
        Qwen2-VL-7B & 32.37 & 71.77 & 54.92 & 83.92 \\
        \hspace{1em} Visual-RFT & 33.81$_{\textcolor{teal}{+1.44}}$ & 57.26$_{\textcolor{red}{-14.51}}$ & 45.90$_{\textcolor{red}{-9.02}}$ & 80.42$_{\textcolor{red}{-3.50}}$ \\
        \hspace{1em} ReasonRFT-ST & 29.50$_{\textcolor{red}{-2.87}}$ & 71.77$_{\textcolor{teal}{+0.00}}$ & 52.46$_{\textcolor{red}{-2.46}}$ & 83.22$_{\textcolor{red}{-0.70}}$ \\
        \hspace{1em} ReasonRFT-SP & 27.34$_{\textcolor{red}{-5.03}}$ & 67.74$_{\textcolor{red}{-4.03}}$ & 54.10$_{\textcolor{red}{-0.82}}$ & 81.35$_{\textcolor{red}{-2.57}}$ \\
        \hspace{1em} ReasonRFT-VC & 31.65$_{\textcolor{red}{-0.72}}$ & 70.97$_{\textcolor{red}{-0.80}}$ & 53.28$_{\textcolor{red}{-1.64}}$ & 82.52$_{\textcolor{red}{-1.40}}$ \\

        \rowcolor{gray!25}
        Qwen2-VL-2B & 33.09 & 56.45 & 44.26 & 65.03 \\
        \hspace{1em} PR1-Counting & 27.34$_{\textcolor{red}{-5.75}}$ & 57.26$_{\textcolor{teal}{+0.81}}$ & 40.16$_{\textcolor{red}{-4.10}}$ & 64.34$_{\textcolor{red}{-0.69}}$ \\
        \hspace{1em} PR1-Grounding & 25.90$_{\textcolor{red}{-7.19}}$ & 53.23$_{\textcolor{red}{-3.22}}$ & 47.54$_{\textcolor{red}{-3.28}}$ & 65.73$_{\textcolor{teal}{+0.70}}$ \\
        \hspace{1em} PR1-OCR & 22.30$_{\textcolor{red}{-10.79}}$ & 57.26$_{\textcolor{teal}{+0.81}}$ & 50.00$_{\textcolor{teal}{+5.74}}$ & 66.43$_{\textcolor{teal}{+1.40}}$ \\
        \hspace{1em} ReasonRFT-VC & 32.37$_{\textcolor{red}{-0.72}}$ & 59.68$_{\textcolor{teal}{+3.23}}$ & 40.16$_{\textcolor{red}{-4.10}}$ & 64.34$_{\textcolor{red}{-0.69}}$ \\

        \rowcolor{gray!25}
        Qwen2.5-VL-3B & 29.89 & 50.81 & 53.28 & 78.32 \\
        \hspace{1em} PR1-Detection & 28.78$_{\textcolor{red}{-1.11}}$ & 61.29$_{\textcolor{teal}{+10.08}}$ & 48.36$_{\textcolor{red}{-4.92}}$ & 80.42$_{\textcolor{teal}{+2.10}}$ \\
        %\hspace{1em} IoU-reward$^\dag$ & - & - & - & - \\

    \hline
        \multicolumn{5}{c}{\textbf{Fine tune on MPCC}} \\
        \rowcolor{gray!10}
        \hspace{1em} prompt     & 29.05$_{\textcolor{red}{-0.84}}$ & 54.84$_{\textcolor{teal}{+4.03}}$ & 55.74$_{\textcolor{teal}{+2.46}}$ & 74.83$_{\textcolor{red}{-5.59}}$ \\
        \hspace{1em} sft        & 34.53$_{\textcolor{teal}{+4.64}}$ & 58.87$_{\textcolor{teal}{+8.06}}$ & 46.72$_{\textcolor{red}{-6.56}}$ & 70.63$_{\textcolor{red}{-7.69}}$ \\
        \hspace{1em} rft        & 36.69$_{\textcolor{teal}{+6.80}}$ & 59.68$_{\textcolor{teal}{+8.87}}$ & 57.38$_{\textcolor{teal}{+4.10}}$ & 77.62$_{\textcolor{red}{-0.70}}$\\
        \hspace{1em} sft+rft    & 31.65$_{\textcolor{teal}{+1.76}}$ & 58.06$_{\textcolor{teal}{+7.25}}$ & 47.54$_{\textcolor{red}{-5.74}}$ & 69.23$_{\textcolor{red}{-9.09}}$\\
        \hspace{1em} rft+$pri.$ & 40.29$_{\textcolor{teal}{+10.40}}$ & 59.68$_{\textcolor{teal}{+8.87}}$ & 57.38$_{\textcolor{teal}{+4.10}}$ & 79.72$_{\textcolor{teal}{+0.90}}$\\
    \hline
    \end{tabular}}
    \caption{Cross-task generalization of strategies. On MPCC task, Qwen2.5-VL-3B as baseline; relative performance changes of each methods shown in table.}
    \label{table4}
    \end{table}

\textbf{Cross-task generalization.} To verify whether the MPCC task effectively activates reasoning, we compared its generalization performance with that of tasks across multiple downstream tasks. We selected four tasks from BLINK \cite{fu2024blink} for evaluation: Spatial Relation (determining the spatial relationship between two objects in an image, Spa.), Semantic Correspondence (identifying the semantically corresponding point between two images, Sem.), Relative Depth (determining which of two given points is closer to the camera, Rel.), and Object Localization (choosing the more accurate bounding box, Obj.).
The methods we compared include Visual-RFT \cite{liu2025visual}, which is based on image classification and object detection tasks, PR1 \cite{yu2025perception} , which skips the reasoning process and uses Counting, Grounding, and Detection for reinforcement fine-tuning, and Reason-RFT \cite{tan2025reason} , which uses Spatial-Transformation (ST), Structure-Perception (SP), and Visual Counting (VC) for reinforcement fine-tuning. 
% In addition, we implemented a model using IoU as the reward signal, under the same data and settings as our method, for performance comparison on the MPCC task.
Table \ref{table4} show that models fine-tuned on the MPCC task achieve significant performance improvements on other tasks, while models fine-tuned on other tasks often lead to performance degradation. For example, Visual-RFT, which is fine-tuned on object detection task, results in a 14.51 performance drop on the Relative Depth task. Although PR1 shows some improvement on certain tasks, it performs worse on those requiring deeper semantic reasoning. Notably, on the relatively simple Spatial Relation task, reasoning may lead to performance degradation, yet our MPCC fine-tuning approach based on RFT with prior sampling still achieves improvement.
Further experiments comparing different fine-tuning strategies on the MPCC task reveal the generalization advantages of reinforcement-based fine-tuning, with the prior sampling strategy achieving the best overall results. Although strategies such as sft or sft+rl can quickly achieve high performance on the training task, they often sacrifice generalization ability. In contrast, RFT with prior sampling strikes a good balance between performance and generalization, further demonstrating the effectiveness of the MPCC task and its fine-tuning strategy in improving model generalization.
More detailed experiments are in the appendix.
%--------------------------------------------------------------------------------------------------------------

\subsection{Case Study and Reasoning analysis}

Figure \ref{fig4} (a) shows several examples of reasoning in the MPCC task. For the first image, our fine-tuned model considers both visual context and commonsense reasoning, listing possible objects like "pot", "spoon", "knife", and "fork", and finally selects the correct one. In the second image, the base model gives a wrong answer directly. Prompt-guided reasoning answers "Bus stop sign", which ignores visual details such as the object being in the middle of the road. The RFT-fine-tuned model answers "bus", which is plausible but ignores mask size. Only the model fine-tuned with SFT+RFT gives the correct answer.
In Figure \ref{fig4} (b), we compare the most frequently used words in the "RFT with prior sampling" and "prompt-guided" reasoning approaches, after removing stop words.  The top 15 words from the fine-tuned model's responses include more context-related terms such as "positioned," "nearby," and "size," as well as expressions indicating commonsense reasoning, like "given," "expect," and "reasonable."  
% This suggests that the model has enhanced capabilities in both visual context understanding and logical inference, contributing to improved generalization performance.
% More case studies and analytical experiments will be presented in the appendix.
This indicates improved capabilities in visual context understanding and logical inference, enhancing generalization performance.  Additional case studies and analyses are provided in the appendix.

\section{Conclusion}

This study introduces MPCC, a new training task to enhance visual understanding and commonsense reasoning in VLMs.    We build the MPCC-Eval benchmark and explore various fine-tuning strategies to guide reasoning development.
Experiments show that MPCC effectively activates multimodal reasoning, especially when combined with RFT with Prior Sampling, achieving strong performance and generalization even with limited data.
Whileour method offers a low-cost path forward, the gap with natural language reasoning remains. Scalable multimodal training with verifiable rewards is still challenging. Future work will focus on reasoning tasks for large-scale training.

\bibliography{main}

\end{document}